\documentclass[11pt,a4paper]{article}
\usepackage[hyperref]{acl2020}
\usepackage{times}
\usepackage{latexsym}

\usepackage{graphicx}

\usepackage{microtype}

\aclfinalcopy % Uncomment this line for the final submission
\def\aclpaperid{2129} %  Enter the acl Paper ID here

\title{A Tale of Two Perplexities: Sensitivity of Neural Language Models to Lexical Retrieval Deficits in Dementia of the Alzheimer's Type}

\author{Trevor Cohen\thanks{$\>$ denotes equal contribution}\\
  Biomedical and Health Informatics\\
  University of Washington \\
  Seattle \\
  \texttt{cohenta@uw.edu} 
  \\\And
  Serguei Pakhomov\footnotemark[1]\\
  Pharmaceutical Care and Health Systems \\
  University of Minnesota \\
  Minneapolis \\
  \texttt{pakh0002@umn.edu} \\}

\date{}

\begin{document}
\maketitle
\begin{abstract}
In recent years there has been a burgeoning interest in the use of computational methods to distinguish between elicited speech samples produced by patients with dementia, and those from healthy controls. The difference between perplexity estimates from two neural language models (LMs) - one trained on transcripts of speech produced by healthy participants and the other trained on transcripts from patients with dementia - as a single feature for diagnostic classification of unseen transcripts has been shown to produce state-of-the-art performance. However, little is known about \textit{why} this approach is effective, and on account of the lack of case/control matching in the most widely-used evaluation set of transcripts (DementiaBank), it is unclear if these approaches are truly diagnostic, or are sensitive to other variables. In this paper, we interrogate neural LMs trained on participants with and without dementia using synthetic narratives previously developed to simulate progressive semantic dementia by manipulating lexical frequency. We find that perplexity of neural LMs is strongly and differentially associated with lexical frequency, and that a mixture model resulting from interpolating control and dementia LMs improves upon the current state-of-the-art for models trained on transcript text exclusively. 
\end{abstract}

%\section{Credits}

%Let's give thanks where thanks are due

\section{Introduction}
Alzheimer's Disease (AD) is a debilitating neurodegenerative condition which currently has no cure, and Dementia of the Alzheimer's Type (DAT) is one of the most prominent manifestations of AD pathology. Prior to availability of disease-modifying therapies, it is important to focus on reducing the emotional and financial burden of this devastating disease on patients, caregivers, and the healthcare system. Recent longitudinal studies of aging show that cognitive manifestations of future dementia may appear as early as 18 years prior to clinical diagnosis - much earlier than previously believed \cite{rajan2015cognitive,aguirre2016cognitive}. With ~30-40\% of healthy adults subjectively reporting forgetfulness on a regular basis \cite{cooper_meaning_2011}, there is an urgent need to develop sensitive and specific, easy-to-use, safe, and cost-effective tools for monitoring AD-specific cognitive markers in individuals concerned about their cognitive function. Lack of clear diagnosis and prognosis, possibly for an extended period of time (i.e., many years), in this situation can produce uncertainty and negatively impact planning of future care \cite{stokes2015dementia}, and misattribution of AD symptoms to personality changes can lead to family conflict and social isolation \cite{boise1999diagnosing,bond2005inequalities}. Delayed diagnosis also results in an estimated \$7.9 trillion in medical and care costs \cite{association_2018_2018} due to high utilization of emergency care, amongst other factors, by patients with undiagnosed AD. 

%contributes to the cost of care of this disease on account of a high utilization of emergency rather than routine care, amongst other factors - it is estimated that early and accurate diagnosis can help save an estimated \$7.9 trillion in medical and care costs \cite{association_2018_2018}. %Furthermore, survey findings show the vast majority (\textasciitilde  80\%) would prefer to know if their unexplained symptoms of confusion or memory loss were due to DAT in a formal clinical evaluation \cite{blendon_five-country_2011}. 

Cognitive status is reflected in spoken language. As manual analysis of such data is prohibitively time-consuming, the development and evaluation of computational methods through which symptoms of AD and other dementias can be identified on the basis of linguistic anomalies observed in transcripts of elicited speech samples have intensified in the last several years
 \cite{fraser2016linguistic, yancheva2016vector, orimaye2017predicting}.  This work has generally employed a supervised machine learning paradigm, in which a model is trained to distinguish between speech samples produced by patients with dementia and those from controls, using a set of deliberately engineered or computationally identified features. However, on account of the limited training data available, overfitting is a concern. This is particularly problematic in DAT, where the nature of linguistic anomalies varies between patients, and with AD progression \cite{altmann2008effects}. %Consequently, a supervised machine learning model trained on one set of patients may not recognize the linguistic anomalies of others. 
 
 In the current study we take a different approach, focusing our attention on the perplexity of a speech sample as estimated by neural LMs trained on transcripts of the speech of participants completing a cognitive task. 
 \iffalse
 In doing so, we confront a paradox of predictability that emerges from prior literature: some authors have found significantly higher perplexity in transcripts from patients with dementia \cite{roark2007}, while others have found perplexity decreases with progression of disease \shortcite{wankerl2016analysis}. Both proposals are plausible. With  elevated perplexity, the underlying idea is that in the context of an elicited speech sample, linguistic anomalies that occur on account of an underlying dementia should appear surprising to a LM trained on text from controls. The diminished perplexity proposal rests on the observation that language use becomes more constrained in DAT in particular, as progressive deficits in semantic memory constrain the vocabulary available for use in speech \cite{altmann2008effects}. 
 \fi
To date, the most successful approach to using LM perplexity as a sole distinguishing feature between narratives by dementia patients and controls was proposed by Fritsch et al. \shortcite{fritsch2019automatic} and replicated by Klumpp et al. \shortcite{klumpp2018}. The approach
consists of training two recurrent neural LMs - one on transcripts from patients with dementia and the other on transcripts from controls. The difference between the perplexities estimated with these two LMs results in very high classification accuracy (AUC: 0.92) reported by both studies. 

The explanation for this performance %why this approach is able to distinguish between DAT and control narratives 
offered by Fritsch et al. \shortcite{fritsch2019automatic} relies on observations that patients with DAT describe the picture in an unforeseen way and their speech frequently diverts from the content of the picture, contains repetitions, incomplete utterances, and refers to objects in the picture using words like ``thing" or ``something". This explanation, however, conflicts with the findings by Klumpp et al. \shortcite{klumpp2018} that demonstrate similarly high classification accuracy (AUC: 0.91) with a single hidden layer non-recurrent neural network and bag-of-words input features, suggesting that while word sequences play a role, it may not be as large as previously believed by Fritsch et al. \shortcite{fritsch2019automatic}.  Klumpp et al.'s \shortcite{klumpp2018} explanation contrasts ``local" with ``global language properties" of the picture descriptions being captured by recurrent neural LMs vs. the non-recurrent bag-of-words neural network classifier, respectively. Both of these explanations are based on informal qualitative observations of the data and are not entirely satisfying because both fail to explain the fact that it is precisely the difference between the control and dementia LMs that is able to discriminate between patients and controls. The individual LMs are not nearly as good at this categorization task. 

%It remains unclear whether these models are truly diagnostic, or are perhaps just sensitive to differences in other subject characteristics such as age or level of education, or some other features of individual language use.  
 
 The objective of the current study is to quantify the extent to which the differences between neural LMs trained on language produced by DAT patients and controls reflect known deficits in language use in this disease - in particular the loss of access to relatively infrequent terms that occurs with disease progression \cite{almor_why_1999}. We approach this objective by interrogating trained neural LMs with two methods:  \textit{interrogation by perturbation} in which we evaluate how trained neural LMs respond to text that has been deliberately perturbed to simulate AD progression; and \textit{interrogation by interpolation} in which we develop and evaluate hybrid LMs by interpolating between neural LMs modeling language use with and without dementia. We find neural LMs are progressively more perplexed by text simulating disease of greater severity, and that this perplexity decreases with increasing contributions of a LM trained on transcripts from patients with AD, but increases again when only this LM is considered. Motivated by these observations, we modify the approach of Fritsch et al. \shortcite{fritsch2019automatic} by incorporating an interpolated model and pre-trained word embeddings, with improvements in performance over the best results reported for models trained on transcript text exclusively.

\section{Background}
\subsection{Linguistic Anomalies in AD}
AD is a progressive disease, and the linguistic impairments that manifest reflect the extent of this progression \cite{altmann2008effects}. In its early stages, deficits in the ability to encode recent memories are most evident. As the disease progresses, it affects regions of the brain that support semantic memory \cite{martin2001semantic} - knowledge of words and the concepts they represent - and deficits in language comprehension and production emerge \cite{altmann2008effects}.  %- particularly where complex phrases and sentences are concerned 
%A motivating argument for the current research rests on the fact that the anatomical spread of the disease varies across individuals, resulting in the categorization of a range of AD subtypes \cite{murray2011neuropathologically}, exacerbating the danger of overfitting supervised machine learning models to a small number of positive examples.

%\subsection{Eliciting Linguistic Anomalies in AD}
A widely-used diagnostic task for elicitation of abnormalities in speech is the ``Cookie Theft'' picture description task from the Boston Diagnostic Aphasia Examination \cite{goodglass2000boston}, which is considered to provide an adequate approximation of spontaneous speech. In this task, participants are asked to describe a picture of a pair of children colluding in the theft of cookies from the top shelf of a raised cupboard while their mother distractedly washes dishes\footnote{For a contemporary edition subscribing to fewer gender stereotypes see \cite{berube2018stealing}.}. When used as a diagnostic instrument, the task can elicit features of AD and other dementias, such as pronoun overuse \cite{almor_why_1999}, repetition \cite{hier_language_1985,pakhomov_recurrent_2018} and impaired recollection of key elements (or ``information units'') from the picture \cite{giles_performance_1996}. Due to the human-intensive nature of the analyses to detect such anomalies, automated methods present a desirable alternative.

\subsection{Classification of Dementia Transcripts}
%examples of prior work using supervised ML
A number of authors have investigated automated methods of identifying linguistic anomalies in dementia. The most widely-used data set for these studies is the DementiaBank corpus \cite{becker1994natural}, which we employ for the current work. In some of the early work on this corpus, Prud'hommeaux and Roark \shortcite{Prudhommeaux2015} introduced a novel graph-based content summary score to distinguish between controls and dementia cases in this corpus with an area under the receiver operating characteristic curve (AUC) of 0.83. Much of the subsequent work relied on supervised machine learning, with a progression from manually engineered features to neural models mirroring general Natural Language Processing trends. For example, Fraser and Hirst \shortcite{fraser2016linguistic} report AD classification accuracy of over 81\% on 10-fold cross-validation when applying logistic regression to 370 text-derived and acoustic features. In a series of papers, Orimaye et al. \shortcite{orimaye2014learning,orimaye2017predicting,orimaye2018deep} report tenfold cross-validation F-measures of up to 0.73 when applying a Support Vector Machine (SVM) to 21 syntactic and lexical features; SVM AUC on leave-pair-out cross-validation (LPOCV) of 0.82 and 0.93 with the best manually-engineered feature set and the best 1,000 of 16,903 lexical, syntactic and n-gram features (with selection based on information gain) respectively; and a LPOCV AUC of 0.73-0.83 across a range of deep neural network models with high-order n-gram features. Yancheva and Rudzicz \shortcite{yancheva2016vector} derive topic-related features from word vector clusters to obtain an F-score of 0.74 with a random forest classifier\footnote{0.8 with additional lexicosyntactic and acoustic features.}. Karlekar et al. \shortcite{karlekar2018detecting} report an utterance-level accuracy of 84.9\%\footnote{This improved to 91.1\% when incorporating POS tags.} with a convolutional/recurrent neural network combination when trained on text alone. While these results are not strictly comparable as they are based on different subsets of the data, use different cross-validation strategies and report different performance metrics, they collectively show that supervised models can learn to identify patients with AD using data from elicited speech samples. However, as is generally the case with supervised learning on small data sets, overfitting is a concern. %to the anomalies of positive cases in the set.  

\subsection{Perplexity and Cognitive Impairment}
Perplexity is used as an estimate of the fit between a probabilistic language model and a segment of previously unseen text. 
%While one would anticipate that perplexity estimates from a LM trained on transcripts from healthy controls would differ in patients with dementia, there is conflicting evidence from prior work as to the nature of these differences. 
%\subsubsection{Elevated Perplexity}
 %prior work using perplexity (Roark, Pakhomov, Wankerl)
The notion of applying n-gram model perplexity (a derivative of cross-entropy) as a surrogate measure of syntactic complexity in spoken narratives was proposed by Roark et al. \shortcite{roark2007} and applied to transcribed logical memory (story recall) test  responses by patients with mild cognitive impairment (MCI:  a frequent precursor to AD diagnosis). In this work, %Roark et al . \shortcite{roark2007} used %manually and automatically derived% 
sequences of part-of-speech (POS) tags were used to train bi-gram models on logical memory narratives, and then cross-entropy of these models was computed on held-out cross-validation folds. They found significantly higher mean cross-entropy values in narratives of MCI patients as compared to controls. Subsequent work expanded the use of POS cross-entropy as one of the language characteristics in a predictive model for detecting MCI \cite{roark2011}. 

Perplexity can also be calculated on word tokens and serve as an indicator of an n-gram model's efficiency in predicting new utterances \cite{jelinek1977}. Pakhomov et al  \shortcite{pakhomov2010computerized} %who 
included word and POS LM perplexity amongst a set of measurements used to distinguish between speech samples elicited from healthy controls and patients with frontotemporal lobar degeneration (FTLD). A LM was trained on text from an external corpus of transcribed ``Cookie Theft'' picture descriptions performed by subjects without dementia from a different study. This model was then used to estimate perplexity of elicited speech samples in cases and controls, with significant differences between mean perplexity scores obtained from subjects with the semantic dementia variant of FTLD and controls. However, the authors did not attempt to use perplexity score as a variable in a diagnostic classification of FTLD or its subtypes.

Collectively, these studies suggest elevated perplexity (both at the word and POS level) may indicate the presence of dementia.
%\subsubsection{Indistinguishable Perplexity}
A follow-up study \cite{pakhomov2010cbn} used perplexity calculated with a model trained on a corpus of conversational speech 
%deleted (Switchboard) to save space 
unrelated to the picture description task, as part of a factor analysis of speech and language characteristics in FTLD. Results suggested that the general English LM word- and POS-level perplexity did not discriminate between FTLD subtypes, or between cases and controls. %say something pertinent about the results 
Taken together with the prior results, these results suggest that LMs trained on transcripts elicited using a defined task (such as the ``Cookie Theft" task) are better equipped to distinguish between cases and controls than LM trained on a broader corpus.

%\subsubsection{Diminished Perplexity}
As the vocabulary of AD patients becomes progressively constrained, one might anticipate language use becoming more predictable with disease progression. Wankerl et al.  \shortcite{wankerl2016analysis} evaluate this hypothesis using the writings of Iris Murdoch who developed AD later in life - and eschewed editorial revisions. In this analysis, which was based on time-delimited train/test splits, perplexity decreased in her later output. This is consistent with recent work by Weiner et al. \shortcite{weiner2018automatic} that found diminished perplexity was of some (albeit modest) utility in predicting transitions to AD.

%\subsection{Two perplexities}
The idea of combining two perplexity estimates - one from a model trained on transcripts of speech produced by healthy controls and the other from a model trained on transcripts from patients with dementia - was developed by Wankerl et al. \shortcite{wankerl2017n} who report an AUC of 0.83 using n-gram LMs in a participant-level leave-one-out-crossvalidation (LOOCV) evaluation across the DementiaBank dataset. Fritsch et al. \shortcite{fritsch2019automatic} further improved performance of this approach by substituting a neural LM (a LSTM model) for the n-gram LM, and report an improved AUC of 0.92. However, it is currently unclear as to whether this level of accuracy is due to dementia-specific linguistic markers, or a result of markers of other significant differences between the case and control group such as age ($\bar{x}$ = 71.4 vs. 63) and years of education ($\bar{x}$= 12.1 vs. 14.3) \cite{becker1994natural}.

%\bigskip
\iffalse
\noindent In summary, while there have been efforts to apply LM perplexity to identify linguistic markers of dementia, there is conflicting evidence of diagnostic utility, particularly in the case of models trained on an on-task corpus. This results in the predictability paradox we attempt to resolve in the current work:  in some cases, perplexity is elevated in the presence of disease, and in others diminished. 
\fi

\subsection{Neural LM perplexity}
Recurrent neural network language models (RNN-LM) \cite{mikolov2010recurrent} are widely used in machine translation and other applications such as sequence labeling \cite{goldberg2016primer}. Recurrent Neural Networks (RNN) \cite{jordan1986serial,elman1990finding} facilitate modeling sequences of indeterminate length by maintaining a \textit{state vector}, $S_{t-1}$, that is combined with a vector representing the input for the next data point in a sequence, $x_t$ at each step of processing. Consequently, RNN-LMs have recourse to information in all words preceding the target for prediction, in contrast to n-gram models. They are also robust to previously unseen word sequences, which with na\"ive n-gram implementations (i.e., without smoothing or backoff) could result in an entire sequence being assigned a probability of zero. Straightforward RNN implementations are vulnerable to the so-called ``vanishing'' and ``exploding'' gradient problems \cite{hochreiter1998vanishing,pascanu2012understanding}, which emerge on account of the numerous sequential multiplication steps that occur with backpropagation through time (time here indicating each step through the sequence to be modeled), and limit the capacity of RNNs to capture long-range dependencies. An effective way to address this problem involves leveraging Long Short Term Memory (LSTM) networks \cite{hochreiter1997long}, which use structures known as gates to inhibit the flow of information during training, and a mechanism using a memory cell to preserve selected information across sequential training steps. Groups of gates comprise vectors with components that have values that are forced to be close to either 1 or 0 (typically accomplished using the sigmoid function). Only values close to 1 permit  transmission of information, which disrupts the sequence of multiplication steps that occurs when backpropagating through time. The three gates used with typical LSTMs are referred to as Input, Forget and Output gates, and as their names suggest they govern the flow of information from the input and past memory to the current memory state, and from the output of each LSTM unit (or cell) to the next training step. LSTM LMs have been shown to produce better perplexity estimates than n-gram models  \cite{sundermeyer2012lstm}.

\subsection{Lexical Frequency}
A known distinguishing feature of the speech of AD patients is that it tends to contain higher frequency words with less specificity than that of cognitively healthy individuals (e.g., overuse of pronouns and words like "thing") \cite{almor1999}. Lexical frequency affects speech production; however, these effects have different origins in healthy and cognitively impaired individuals. A leading cognitive theory of speech production postulates a two-step process of lexical access in which concepts are first mapped to lemmas and, subsequently, to phonological representations prior to articulation \cite{Levelt2001}. In individuals without dementia, lexical frequency effects are evident only at the second step - the translation of lemmas to phonological representations and do not originate at the pre-lexical conceptual level \cite{jescheniak_word_1994}. In contrast, in individuals with dementia, worsening word-finding difficulties are attributed to progressive degradation of semantic networks that underlie lexical access at the conceptual level \cite{ASTELL1996}. While lexical frequency effects are difficult to control in unconstrained purely spontaneous language production, language produced during the picture description task is much more constrained in that the picture provides a fixed set of objects, attributes, and relations that serve as referents for the the person describing the picture. Thus, in the context of the current study, we expect to find that both healthy individuals and patients with dementia describing the same picture would attempt to refer to the same set of concepts, but that patients with dementia would tend to use more frequent and less specific words due to erosion of semantic representations leading to insufficient activation of the lemmas. Changes in vocabulary have been reported in the literature as one of the most prominent linguistic manifestations of AD {\cite{pekkala2013,wilson1983,rohrer2007}}. We do not suggest that other aspects of language such as syntactic complexity, for example, should be excluded; although, there has been some debate as to the utility of syntactic complexity specifically as a distinguishing feature (see {\cite{fraser2015}).

\section{Materials and Methods}
\subsection{Datasets} 
For \textit{LM training and evaluation} we used transcripts of English language responses to the ``Cookie Theft''  component of the Boston Diagnostic Aphasia Exam \cite{goodglass2000boston}, provided as part of the  DementiaBank database \cite{becker1994natural}. Transcripts (often multiple) are available for 169 subjects classified as having possible or probable DAT on the basis of clinical or pathological examination, and 99 patients classified as controls.  %Of these 99, 10 patients subsequently received a dementia-related diagnosis, according to the metadata made available along with the DementiaBank database. According to the metadata, 7 of these 10 patients had a probable AD diagnosis and 3 had an indeterminate diagnostic status at baseline. Thus, the remaining 89 of the 99 patients represent the set of controls that are less likely to have had any underlying AD pathology present but undetected at baseline. In addition, only 148 of the dementia cases are assigned the category ``probable AD'' which indicates a high level of diagnostic suspicion of AD. To maintain consistence with prior work \cite{orimaye2017predicting,orimaye2018deep}, we restrict our evaluations to these 148 individuals. 

%Participants:  99  controls and  169  cases %Transcripts :  242  controls and  257  cases 
For \textit{interrogation by perturbation}, we used a set of six synthetic ``Cookie Theft'' picture description narratives created by Bird et al. \shortcite{bird2000} to study the impact of semantic dementia on verb and noun use in picture description tasks. While Bird et al. \shortcite{bird2000} focused on semantic dementia, a distinct condition from DAT, these synthetic narratives were not based on patients with semantic dementia. Rather, they were created to manipulate lexical frequency by first compiling a composite baseline narrative from samples by healthy subjects, and then removing and/or replacing nouns and verbs in that baseline with words of higher lexical frequency (e.g., ``mother'' vs. ``woman'' vs. ``she''). Lexical frequency was calculated using the Celex Lexical Database (LDC96L14) and words were aggregated into groups based on four log frequency bands (0.5 - 1.0, 1.0 - 1.5, 1.5 - 2.0, 2.5 - 3.0: e.g., words in the 0.5 - 1.0 band  occur in Celex more than 10 times per million). These narratives are well-suited to the study of lexical retrieval deficits in DAT in which loss of access to less frequent words is observed with disease progression \cite{pekkala2013}.      

In order to \textit{calculate mean log lexical frequency on the DementiaBank narratives}, we used the SUBTLEX\textsubscript{us} corpus shown to produce lexical frequencies more consistent with psycholinguistic measures of word processing time than those calculated from the Celex corpus \cite{Brysbaert2009}. The DementiaBank narratives were processed using NLTK's \footnote{Natural Language Toolkit: www.nltk.org} implementation of the TnT part-of-speech tagger \cite{brants2000tnt} trained on the Brown corpus \cite{francis79browncorpus}. Following Bird et al. \shortcite{bird2000} only nouns and verbs unique within the narrative were used to calculate mean log lexical frequency. We did not stem the words in order to avoid creating potentially artificially high/low frequency items. To validate the mean log lexical frequency values obtained with the SUBTLEX\textsubscript{us} corpus, we compared the log lexical frequency means for the six narratives developed by Bird et al. \shortcite{bird2000} with their frequency band values using Spearman's rank correlation and found them to be perfectly correlated ($\rho$ = 1.0). 

The text of DementiaBank transcripts was extracted from the original CHAT files \cite{Macwhinney2000}. The transcripts as well as the six synthetic narratives were lowercased and pre-processed by removing speech and non-speech noise as well as pause fillers (um's amd ah's) and punctuation (excepting the apostrophe). 

\subsection{Pre-trained models}
Prior work with neural LMs in this context has used randomly instantiated models. We wished to evaluate the utility of pre-training for this task - both pre-training of the LSTM in its entirety and pre-training of word embeddings alone. For the former we used a LSTM trained on the WikiText-2 dataset \cite{merity2016pointer} provided with the \texttt{GluonNLP} package\footnote{https://github.com/dmlc/gluon-nlp}.  
200-dimensional word embeddings, including embeddings augmented with subword information,  \cite{bojanowski2017enriching} %and positional \cite{cohen2018bringing}
were developed using the \texttt{Semantic Vectors} package\footnote{https://github.com/semanticvectors/semanticvectors} and trained using the skipgram-with-negative-sampling algorithm of Mikolov et al.   \shortcite{mikolov2013distributed} for a single iteration on the English Wikipedia (10/1/2019 edition, pre-processed with \texttt{wikifl.pl}\footnote{Available at  https://github.com/facebookresearch/fastText}) with a window radius of five\footnote{Other hyperparameters per \cite{cohen2018bringing}}. We report results using skipgram embeddings augmented with subword information as these improved performance over both stochastically-initialized and WikiText-2-pretrained LSTMs in preliminary experiments.

%In addition, we evaluated 200-dimensional \texttt{Glove} \cite{pennington2014glove} embeddings \footnote{Available at https://nlp.stanford.edu/projects/glove/} trained on the Gigaword 5 corpus\footnote{https://catalog.ldc.upenn.edu/LDC2011T07} and a 2014 release of Wikipedia. 
%todo - rewrite below

\subsection{Training}
\iffalse
N-gram LMs (4-grams) were developed using the SRI Language Modeling Toolkit (SRILM) \shortcite{Stolcke2002SRILMA}, with Kneser-Ney smoothing \cite{ney1994} and frequency cutoffs of 2 for unigrams and bigrams and 1 for higher order n-grams. Out of vocabulary words and sentence boundaries were ignored during perplexity calculations. 
\fi
We trained two sets of dementia and control LSTM models. The first set was trained in order to replicate the findings of Fritsch et al. \shortcite{fritsch2019automatic}, using the same \texttt{RWTHLM} package \cite{sundermeyer2014} and following their methods as closely as possible in accordance with the description provided in their paper. Each model's cross-entropy loss was optimized over 20 epochs with starting learning rate optimization performed on a heldout set of 10 transcripts. The second set was trained using the \texttt{GluonNLP} averaged stochastic gradient weight-dropped LSTM (standard-lstm-lm-200 architecture) model consisting of 2 LSTM layers with word embedding (tied at input and output) and hidden layers of 200 and 800 dimensions respectively (see Merity et al. \shortcite{merity2017regularizing} for full details on model architecture). In training the \texttt{GluonNLP} models, the main departure from the methods used by Fritsch et al. \shortcite{fritsch2019automatic} involved not using a small heldout set of transcripts to optimize the learning rate because we observed that the \texttt{GluonNLP} models converged well prior to the 20th epoch with a starting learning rate of 20 which was used for all stochastically initialized models. With pre-trained models we used a lower starting learning rate of 5 to preserve information during subsequent training on DementiaBank. All \texttt{GluonNLP} models were trained using batch size of 20 and back propagation through time (BPTT) window size of 10. During testing, batch size was set to 1 and BPTT to the length of the transcript (tokens). Unseen transcript perplexity was calculated as $e^{\mathrm{loss}}$.

\iffalse
In addition to LSTM and n-gram methods, we experimented with variants of SG and EARP models including directional, permutation and proximity models with and without pre-trained subword embeddings... need a better description here.
\fi
\subsection{Evaluation}
As subjects in the DementiaBank dataset participated in multiple assessments, there are multiple transcripts for most of the subjects. In order to avoid biasing the models to individual subjects, we followed the participant-level leave-one-out cross-validation (LOOCV) evaluation protocol of Fritsch et al. \shortcite{fritsch2019automatic} whereby all of the picture description transcripts for one participant are held out in turn for testing and the LMs are trained on the remaining transcripts. Perplexities of the LMs are then obtained on the heldout transcripts, resulting in two perplexity values per transcript, one from the LM trained on the dementia ($P_{dem}$) and control ($P_{con}$) transcripts. Held-out transcripts were scored using these perplexity values, as well as by the difference ($P_{con} - P_{dem}$) %and ratio ($\frac{P_{con}}{P_{dem}}$) 
between them.
\iffalse
Because subjects in the DementiaBank dataset participated in multiple assessments, there are multiple transcript for most of the subjects. In order to avoid biasing the models to individual subjects, at each LOOCV iteration, in addition to the one transcript heldout for testing, we also withheld from training all other transcripts belonging to the same subject.  
\fi
\begin{table*}
\begin{tabular}{lllllll}
& \multicolumn{2}{c}{CONTROL} & \multicolumn{2}{c}{DEMENTIA} & \multicolumn{2}{c}{CONTROL-DEMENTIA} \\
MODEL & AUC  & 95\% CI & AUC  & 95\% CI & AUC  & 95\% CI \\
\hline
%$SRILM_{Ngram}$ & 0.77 & -- & 0.63 & -- & 0.86 & --  \\
$\texttt{RWTHLM}_{LSTM}$  & 0.80 & -- & 0.64 & -- & 0.92  & --  \\
$\texttt{GluonNLP}_{LSTM}$  & 0.80 & $\pm$ 0.002 & 0.65 & $\pm$ 0.002 & 0.91 & $\pm$ 0.004  \\
\end{tabular}
\caption{Classification accuracy using individual models' perplexities and their difference for various  models.}
\label{pplModelsTable}
\end{table*}
\subsection{Interrogation of models}
For \textit{interrogation by perturbation}, we estimated the perplexity of our models for each of the six synthetic narratives of Bird et al. \shortcite{bird2000}. We reasoned that an increase in $P_{con}$ and a decrease in  $P_{dem}$ as words are replaced by higher-frequency alternatives to simulate progressive lexical retrieval deficits would indicate that these models were indeed capturing AD-related linguistic changes. For \textit{interrogation by interpolation}, we extracted the parameters from all layers of paired LSTM LMs after training, and averaged these  as $\alpha LM_{dem} + (1-\alpha) LM_{con}$ to create interpolated models. We hypothesized that a decrease in perplexity estimates for narratives emulating severe dementia would occur as $\alpha$ (the proportional contribution of $LM_{dem}$) increases.

%\subsection{Statistical Methods}
%To compare LM perplexity values with lexical frequency values, we fitted linear regression models with mean lexical frequency as the dependent variable and the dementia and control model perplexities as independent variables. Models were adjusted for age, education and the length of the picture description narrative. 
%A regression model with mean lexical frequency as the dependent variable and only age, sex, education and narrative length as covariates showed that there was no significant association between sex and lexical frequency; therefore, sex was not included as a covariate in subsequent modeling. 
%In order to avoid likely practice effects across multiple transcripts, we only used the transcript obtained on the initial baseline visit;however, we did confirm these results by using all transcripts to fit mixed effects models with random slopes and intercepts in order to account for the correlation between transcripts from the same subject (mixed effects modeling results not shown). An alpha threshold of 0.05 was used to determine statistical significance. 

\section{Results and Discussion}
The results of evaluating classification accuracy of the various language models are summarized in Table \ref{pplModelsTable}. The 95\% confidence interval for \texttt{GluonNLP} models was calculated from perplexity means obtained across ten LOOCV iterations with random model weight initialization on each iteration. The \texttt{RWTHLM} package does not provide support for GPU acceleration and requires a long time to perform a single LOOCV iteration (approximately 10 days in our case). Since the purpose of using the \texttt{RWTHLM} package was to replicate the results previously reported by Fritsch et al. \shortcite{fritsch2019automatic} that were based on a single LOOCV iteration and we obtained the exact same AUC of 0.92 on our first LOOCV iteration with this approach, we did not pursue additional LOOCV iterations. However, we should note that we obtained an AUC of 0.92 for the difference between $P_{con}$ and $P_{dem}$ on two of the ten LOOCV iterations with the \texttt{GluonNLP} LSTM model. Thus, we believe that the \texttt{GluonNLP} LSTM model has equivalent performance to the \texttt{RWTHLM} LSTM model. %The AUC for the difference in Control and Dementia models' perplexities with the n-gram approach is significantly worse than the AUCs for eaithe rof the LSTM models. The AUC of 0.86 obtained with the SRILM n-gram models are consistent with previously reported AUC of 0.83 by Wankerl et al. \shortcite{wankerl2017n}. 
%results to pick embeddings (with slightly different configuration - no unification of vocabularies across models - and hence slightly lower results for baseline and overall

\iffalse
With AUCs above 0.90 for most of the models (except n-gram), the results shown in Table \ref{pplModelsTable} exceed previously reported results for supervised models trained on a range of manually engineered feature sets \cite{orimaye2017predicting}, and assorted deep neural network configurations \cite{orimaye2018deep}. Better AUCs (above 0.90) were obtained by Orimaye et al. \shortcite{orimaye2017predicting} with models trained on the best 1,000 of around 17,000 features, and it seems likely that the performance of our perplexity estimates alone would be surpassed if they were included amongst a broader range of features for supervised learning. However, following the rule of thumb of 10 samples per feature per category in regression modeling, with only 198 samples (99 cases and controls), it is likely that predictive feature sets significantly larger than 10 variables will result in overfitting and reduced chances for generalization to external data. More parsimonious models with fewer features are likely to be more generalizable offer better interpretability.
\fi
Having replicated results of previously published studies and confirmed that using the difference in perplexities trained on narratives by controls and dementia patients is indeed the current state-of-the-art, we now turn to explaining why the difference between these LMs is much more successful than the individual models alone.  
\begin{figure}[t]
  \includegraphics[width=\linewidth]{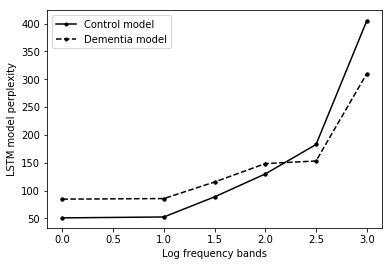} 
  \caption{Relationship between log frequency bands used to replace words in synthetic Cookie Theft picture descriptions to simulate degrees of semantic dementia and perplexity of LSTM language models trained on picture descriptions by controls and dementia patients.}
  \label{fig:ttp}
\end{figure}

First, we used the six ``Cookie Theft'' narratives designed to simulate semantic dementia to examine the relationship between $P_{con}$ and $P_{dem}$ with \texttt{GluonNLP} LSTM LMs and log lexical frequency bands. The results of this analysis are illustrated in Figure \ref{fig:ttp} and show that $P_{dem}$ is higher than $P_{con}$ on narratives in the lower log frequency bands (less simulated impairment) and lower in the higher log frequency bands (more simulated impairment). 

We confirmed these results by calculating mean log lexical frequency on all DementiaBank narratives and fitting a linear regression model to test for associations with perplexities of the two LMs. The regression model contained mean lexical frequency as the dependent variable and $P_{dem}$ and $P_{con}$  as independent variables, adjusted for age, education and the length of the picture description narrative. In order to avoid likely practice effects across multiple transcripts, we only used the transcript obtained on the initial baseline visit;
however, we did confirm these results by using all transcripts to fit mixed effects models with random slopes and intercepts in order to account for the correlation between transcripts from the same subject (mixed effects modeling results not shown).

The results demonstrate that the association between perplexity and lexical frequency is significant and positive for the control LM (coeff: 0.563, p $<$ 0.001) and negative for dementia LM (coeff: -0.543, p $<$ 0.001). Age, years of education, and length of the narrative were not significantly associated with lexical frequency in this model. These associations show that the control LM and dementia LM are more ``surprised" by narratives containing words of higher lexical frequency and lower lexical frequency respectively. If the use of higher lexical frequency items on a picture description task portends a semantic deficit, then this particular pattern of results explains why it is the difference between the two models that is most sensitive to manifestations of dementia and suggests that there is a point at which the two models become equally ``surprised" with a difference between their perplexities close to zero. In Figure \ref{fig:ttp}, that point is between log lexical frequency bands of 2.0 and 2.5 corresponding to the mild to moderate degree of semantic impairment reported by Bird et al. \shortcite{bird2000}. Notably, in the clinical setting, the mild forms of dementia such as mild cognitive impairment and mild dementia are also particularly challenging and require integration of multiple sources of evidence for accurate diagnosis \cite{KNOPMAN2014}.

\begin{table*}
\begin{tabular}{lrrrrrrrrr}
& \multicolumn{2}{c}{RANDOM}  & \multicolumn{2}{c}{PRETRAINED} & \multicolumn{2}{c}{RANDOM}  & \multicolumn{2}{c}{PRETRAINED} \\
$P_{con} - P_\alpha$ &  $AUC$& $95\%\ CI$ & $AUC$  &  $95\%\ CI$ &  $ACC_{eer}$& $95\%\ CI$ & $ACC_{eer}$  & $95\%\ CI$ CI \\
\hline
$\alpha=0.25$   &  0.842 &  $\pm$ 0.008 &  0.838 &  $\pm$ 0.015 &  0.689 &  $\pm$ 0.036 &  0.724 &  $\pm$ 0.034 \\
$\alpha=0.5$    &  0.816 &  $\pm$ 0.009  &  0.813 &  $\pm$ 0.005 &  0.669 &  $\pm$ 0.035  &  0.665 &  $\pm$ 0.033 \\
$\alpha=0.75$   &  \textbf{0.931} &  \textbf{$\pm$ 0.003} &  \textbf{0.941} &  \textbf{$\pm$ 0.006} &  \textbf{0.854} &  \textbf{$\pm$ 0.031} &  \textbf{0.872} &  \textbf{$\pm$ 0.010} \\
\hline
$\alpha=1.0$   &  \textit{0.908} &  $\pm$ \textit{0.004} &  0.930 &  $\pm$ 0.005 &  \textit{0.846} &  $\pm$ \textit{0.023} &  0.839 &  $\pm$ 0.017 \\
\end{tabular}
\caption{Performance of randomly-instantiated and pre-trained (subword-based skipgram embeddings) interpolated ``two perplexity'' models across 10 repeated per-participant LOOCV runs. $\alpha$ indicates the proportional contribution of the dementia model. $ACC_{eer}$ gives the accuracy at equal error rate. Best results are in \textbf{boldface}, and results using the approach of Fritsch et al. \shortcite{fritsch2019automatic} are in \textit{italics}. }
\label{interpolatedModelsTable}
\end{table*}

The results of our interpolation studies are shown in Figure \ref{fig:mixed_bird}. Each point in the figure shows the average difference between the perplexity estimate of a perturbed transcript ($Px$) and the perplexity estimate for the unperturbed ($Po$: frequency band 0) sample for this model\footnote{We visualized this difference because perplexities at $\alpha$=0.5 were generally higher, irrespective of whether component models were initialized stochastically, or had pre-trained word embeddings in common. Perplexities of $\alpha$=0.75 models were slightly lower than those of their majority constituents.}. While all models tend to find the increasingly perturbed transcripts more perplexing than their minimally perturbed counterparts, this perplexity decreases with increasing contributions of the dementia LM. However, when only this model is used, relative perplexity of the perturbed transcripts increases. This indicates that the ``pure'' dementia LM may be responding to linguistic anomalies other than those reflecting lack of access to infrequently occurring terms. We reasoned that on account of this, the $\alpha$=0.75 model may provide a better representation of dementia-related linguistic changes. To evaluate this hypothesis, we assessed the effects on performance of replacing the dementia model with this interpolated model. The results of these experiments (Table \ref{interpolatedModelsTable}) reveal improvements in performance with this approach, with best AUC (0.941) and accuracy at equal error rate (0.872) resulting from the combination of interpolation\footnote{Simply weighting the difference in model perplexities does not perform as well as interpolating model weights, with at best a 0.001 improvement in AUC over the baseline.} with pre-trained word embeddings. That pre-trained embeddings further improve performance is consistent with the observation that the elevation in perplexity when transitioning from $\alpha$=0.75 to $\alpha$=1.0 is much less pronounced in these models (Figure \ref{fig:mixed_bird_pt}). These results are significantly better than those reported by Fritsch et al \shortcite{fritsch2019automatic}, and our reimplementation of their approach. 

\begin{figure}
  \includegraphics[width=\linewidth]{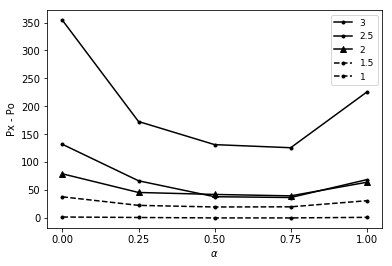} 
  \caption{Stochastically initialized models. Elevation in perplexity over unperturbed transcript ($Po$) with the proportional contribution of a dementia model ($\alpha$) to an interpolated model. Each point is the mean of 268 (held-out participants) data points. Error bars are not shown as they do not exceed the bounds of the markers. }
  \label{fig:mixed_bird}
\end{figure}

\begin{figure}
  \includegraphics[width=\linewidth]{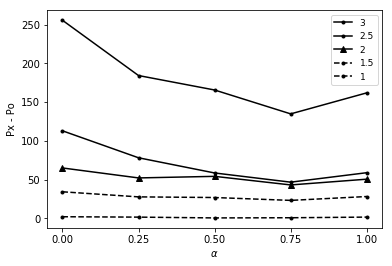} 
  \caption{Pretrained word embeddings. Elevation in perplexity over unperturbed transcript ($Po$) with the proportional contribution of a dementia model ($\alpha$) to an interpolated model. Each point is the average of 268 data points, and error bars are not shown as they do not exceed the bounds of the markers. }
  \label{fig:mixed_bird_pt}
\end{figure}

These improvements in performance appear to be attributable to a smoothing effect on the perplexity of the modified dementia models in response to unseen dementia cases. Over ten repeated LOOCV iterations, average perplexity on held-out dementia cases was significantly lower than that of the baseline `dementia' model (51.1 $\pm$0.81) for both the $\alpha$=0.75  (47.3$\pm$0.32) and pre-trained embeddings (44.8$\pm$0.53) models. This trend is further accentuated with the severity of dementia - for transcripts corresponding to a mini-mental state exam  (MMSE) $\leq$ 10 (n=16), average perplexities are 148.29$\pm$7.69, 105.01$\pm$3.48 and 121.86$\pm$7.67 for baseline `dementia', $\alpha$=0.75 and pre-trained embeddings models respectively. In both cases, average perplexity of the interpolated ($\alpha$=0.75) pre-trained embeddings model fell between those of the exclusively pre-trained (lowest overall) and exclusively interpolated (lowest in severe cases) models. 

A practical issue for automated methods to detect dementia concerns establishing their accuracy at earlier stages of disease progression, where a readily disseminable screening tool would arguably have greatest clinical utility, especially in the presence of an effective disease-modifying therapy. To this end, Fritsch et al. \shortcite{fritsch2019automatic} defined a ``screening scenario'' in which evaluation was limited to participants with a last available MMSE of 21 or more, which corresponds to a range of severity encompassing mild, questionable or absent dementia \cite{perneczky2006mapping}. In this scenario, classification accuracy of the ‘paired perplexity’ LSTM based model was only slightly lower (AUC: 0.87) than the accuracy on the full range of cognitive impairment (AUC: 0.92). We found similar performance with our models. When limiting evaluation to those participants with a last-recorded MMSE $\geq$  21, average AUCs across 10 LOOCV iterations were 0.836 $\pm$0.014, 0.879 $\pm$0.01, 0.893 $\pm$0.004, and 0.899 $\pm$0.012 for the baseline (Fritsch et al \shortcite{fritsch2019automatic}), pretrained embeddings, interpolated ($\alpha$=0.75) and interpolated ($\alpha$=0.75) with pretrained embeddings variants, respectively. These results support the notion that paired neural LMs can be used effectively to screen for possible dementia at earlier stages of cognitive impairment. 

The contributions of our work can be summarized as follows. First, our results demonstrate that the relationship between LM perplexity and lexical frequency is consistent with the phenomenology of DAT and its deleterious effects on patients' vocabulary. We show that the ``two perplexities'' approach is successful at distinguishing between cases and controls in the DementiaBank corpus \textit{because of} its ability to capture specifically linguistic manifestations of the disease.
Second, we observe that interpolating between dementia and control LMs mitigates the tendency of dementia-based LMs to be ``surprised'' by transcripts indicating severe dementia, which is detrimental to performance when the difference between these LMs is used as a basis for classification. In addition, we find a similar smoothing effect when using pre-trained word embeddings in place of a randomly instantiated word embedding layer.
Finally, we develop a modification of Fritsch et al's ``two perplexity'' approach that is consistent with these observations - replacing the dementia model with an interpolated variant, and introducing pre-trained word embeddings at the embedding layer. Both modifications exhibit significant improvements in performance, with best results obtained by using them in tandem. Though not strictly comparable on account of differences in segmentation of the corpus amongst others, we note the performance obtained also exceeds that reported with models trained on text alone in prior research. Code to reproduce the results of our experiments is available on GitHub\footnote{https://github.com/treversec/tale\_of\_two\_perplexities}.

While using transcript text directly is appealing in its simplicity, others have reported substantial improvements in performance when POS tags and paralinguistic features are incorporated, suggesting fruitful directions for future research. Furthermore, prior work on using acoustic features shows that they can contribute to discriminative models \cite{konig2015}; however, Dementia Bank audio is challenging for acoustic analysis due to poor quality and background noise. Lastly, while our results do support the claim that classification occurs on the basis of dementia-specific linguistic anomalies, we also acknowledge that DementiaBank remains a relatively small corpus by machine learning standards, and that more robust validation would require additional datasets.

\section{Conclusion}
We offer an empirical explanation for the success of the difference between neural LM perplexities in discriminating between DAT patients and controls, involving lexical frequency effects. Interrogation of control- and dementia-based LMs using synthetic transcripts and interpolation of parameters reveals inconsistencies harmful to model performance that can be remediated by incorporating interpolated models and pre-trained embeddings, with significant performance improvements.

% In addition 

\iffalse
Our study has several limitations that should be considered in the interpretation of the results and future work. While DementiaBank is invaluable as a publicly available resource to facilitate comparative evaluation of computational diagnostic methods, patients and controls in this dataset were not matched on age (a major risk factor for AD) and education (associated with results of language based neurocognitive tests). Therefore, one must exercise caution in working with these data to ensure that these variables are taken into account as potential confounders. While DementiaBank is currently the largest publicly available and well-characterized dataset of picture descriptions by patients with DAT and controls, it is still relatively small. It is necessary to validate the findings reported in this paper using other datasets to determine the degree of generalizability of neural LMs trained on or finetuned to DementiaBank data. 
\fi
%acc at eer calculated as: %reslist.append(tpr[np.nanargmax(np.absolute((tpr - fpr)))])

%TODO - add pretrained

\section*{Acknowledgments}
This research was supported by Administrative Supplement R01 LM011563 S1 from the National Library of Medicine.
%The acknowledgments should go immediately before the references.  Do
%not number the acknowledgments section. Do not include this section
%when submitting your paper for review. \\

%\noindent {\bf Preparing References:} \\
%Include your own bib file like this:
%\verb|\bibliographystyle{acl_natbib}|
%\verb|\bibliography{naaclhlt2019}| 

%where \verb|naaclhlt2019| corresponds to a naaclhlt2019.bib file.
\bibliographystyle{acl_natbib}
\bibliography{acl2020.bib}

\end{document}